\documentclass[a4paper,fleqn]{cas-sc}

\usepackage[authoryear,longnamesfirst]{natbib}

\usepackage{amsmath,amsfonts}
\usepackage{calrsfs}
\usepackage{algorithmic}
\usepackage{algorithm}
\usepackage{array}
\usepackage{graphicx}
\usepackage{xcolor}
\usepackage{booktabs}
\usepackage{multirow}
\usepackage{tikz}
\usepackage{stfloats}
\usepackage{url}
\usepackage{verbatim}
\usepackage{textcomp}
\usepackage{subfig}

\begin{document}
\let\WriteBookmarks\relax
\def\floatpagepagefraction{1}
\def\textpagefraction{.001}

\shorttitle{Interpretable Prompt Learning}
\shortauthors{Wang et~al.}

\title[mode=title]{Joint Semantic Token Selection and Prompt Optimization for Interpretable Prompt Learning}
% \tnotetext[t1]{Preprint. Under review.}

% ===== Authors =====

\author[inst1]{Yating Wang}

\author[inst1]{Yaqi Zhao}

\author[inst1]{Yongshun Gong}

\author[inst1]{Yilong Yin}

\author[inst1]{Haoliang Sun}
\cormark[1]
\ead{haolsun@sdu.edu.cn}

\cortext[cor1]{Corresponding author}

% ===== Affiliation =====

\affiliation[inst1]{
    organization={School of Software, Shandong University},
    city={Jinan},
    postcode={250101},
    country={China}
}

\begin{abstract}
Vision-language models such as CLIP achieve strong visual-textual alignment, but often suffer from overfitting and limited interpretability when adapted through continuous prompt learning. While discrete prompt optimization improves interpretability, it usually depends on large external models, leading to high computational costs and limited scalability. In this paper, we propose Interpretable Prompt Learning (IPL), a hybrid framework that alternates between discrete semantic token selection and continuous prompt optimization. Specifically, IPL formulates semantic token selection as an approximate submodular optimization problem, encouraging tokens that are both human-understandable and semantically diverse. It further adopts an alternating optimization strategy to integrate discrete token selection with continuous prompt tuning, improving interpretability while preserving adaptability to downstream tasks. Our framework is plug-and-play, allowing seamless integration with existing prompt learning methods. Extensive experiments on multiple benchmarks show that IPL consistently improves both interpretability and accuracy across five representative prompt learning methods, providing an effective and scalable extension to existing frameworks. 
\end{abstract}


\maketitle

\section{Introduction}
\label{sec:introduction}

Vision-Language Models (VLMs) \citep{alayrac2022flamingo, jia2021scaling, radford2021learning, yao2021filip, yu2022coca, yuan2021florence, zhai2022lit, li2022blip} have made remarkable progress in recent years by leveraging large-scale datasets to jointly learn image and text representations. Their exceptional versatility stems from the ability to align visual and textual modalities within a shared representation space, enabling strong zero-shot generalization across diverse vision-language scenarios. However, their substantial model size and complexity often pose significant challenges for adaptation to specific downstream tasks \citep{zhou2022learning, zhou2022conditional, sung2022vl}.

To further improve the adaptability of VLMs to downstream tasks with limited labeled data, prompt learning \citep{jia2022visual, zhou2022learning, zhou2022conditional} has become a widely adopted approach. It typically fine-tunes a small set of learnable prompt tokens while keeping the backbone model frozen, enabling task adaptation with minimal computational overhead. These continuously optimized methods have demonstrated impressive performance, even in few-shot scenarios. Despite their success, these methods face notable challenges. One major issue is overfitting, where learned prompts become overly specialized to the training set, leading to poor generalization and reduced interpretability \citep{ma2023understanding, park2024prompt}. Moreover, because these prompt tokens are optimized in a continuous embedding space, they often drift away from interpretable semantic tokens, making the learned knowledge harder to interpret or transfer. In addition, several existing methods rely heavily on external models, such as large multimodal or language models, to generate or refine prompts \citep{li2025advancing, du2024ipo}. The dependence on such external models also introduce potential instability and restricts their flexibility in real-world applications, particularly when dealing with long or complex textual contexts \citep{liu2023lost, li2025longcontext}.

Building on these observations, we propose Interpretable Prompt Learning (IPL), a method that combines discrete semantic token selection with continuous prompt optimization. In the discrete stage, we adopt a greedy selection strategy formulated as an approximate submodular optimization problem to identify a set of tokens from a candidate pool, ensuring semantic alignment with the target domain. These semantic tokens serve as interpretable pivots that capture meaningful visual and textual concepts, such as object attributes or physical traits, thereby providing human-understandable semantics for prompt construction. In the continuous stage, we fine-tune the learnable embeddings while keeping the semantic tokens fixed, allowing soft tokens to adapt to the target dataset without losing their semantic grounding. By alternating between these two stages, IPL unifies discrete semantic selection and continuous task adaptation within a single framework. Unlike prior methods that depend on external large models, IPL performs both semantic discovery and optimization internally, enabling efficient and scalable deployment.

Beyond the architectural design, semantic tokens provide a principled mechanism for improving prompt learning. Because these tokens correspond to explicit and transferable concepts, the learned prompts can capture features that better reflect the underlying structure of the target domain, thereby enhancing interpretability and transparency. Moreover, fixing part of the prompt as semantic tokens, rather than treating all prompt elements as fully learnable embeddings, reduces the number of free parameters and introduces useful semantic priors. This design helps mitigate overfitting and improves generalization to unseen categories. It also implicitly encourages consistency between visual and textual modalities, leading to more stable transfer across different tasks.

Our approach offers a simple yet effective enhancement to existing prompt learning methods and can be seamlessly integrated with various prompt learning frameworks, making it a versatile and broadly applicable solution. In summary, our contributions are as follows:

\begin{itemize}
\item We propose a novel framework that alternates between semantic token selection and continuous prompt optimization, achieving a balance between prompt interpretability and adaptability.

\item We formulate semantic selection as an approximate submodular optimization problem and design a submodular function that combines cross-entropy with a diversity penalty, encouraging prompt diversity and improving generalization to unseen categories.

\item We demonstrate the effectiveness of our approach by integrating it into five widely used prompt learning methods, achieving consistent gains in both interpretability and performance across diverse benchmarks.

\end{itemize}

\section{Related Works}

\subsection{Vision-Language Models}
Vision-language models (VLMs) have achieved remarkable progress by learning aligned visual and textual representations from large-scale image--text pairs, enabling strong multimodal understanding and transferability. Representative models, such as CLIP \citep{radford2021learning}, ALIGN \citep{jia2021scaling}, BLIP \citep{li2022blip}, LiT \citep{zhai2022lit}, and Flamingo \citep{alayrac2022flamingo}, adopt different pretraining strategies to bridge vision and language. Among them, CLIP has become one of the most influential foundations for downstream adaptation, as it maps images and texts into a shared embedding space through contrastive learning and supports strong zero-shot recognition.

Recent VLMs have further expanded this paradigm by incorporating larger multimodal architectures and stronger reasoning capabilities, such as BLIP-2 \citep{li2023blip} and LLaVA-based models \citep{liu2024llava}. At the same time, recent surveys have shown that VLM research has increasingly shifted from pretraining alone toward efficient adaptation strategies, including fine-tuning, prompt engineering, and parameter-efficient transfer \citep{danish2025comprehensive,shinde2025survey}. This trend highlights that, despite strong pretrained alignment, effectively adapting VLMs to downstream tasks remains a central challenge.

In particular, CLIP-style models often require task-specific adaptation to achieve stronger semantic alignment and transferability on downstream benchmarks. This has motivated the development of prompt learning methods, which adapt pretrained VLMs by optimizing only a small set of prompt parameters while keeping the backbone frozen. Recent overviews of CLIP prompting further indicate that improving generalization and maintaining interpretability have become key concerns in this line of research \citep{cui2025generalizable}.

\subsection{Prompt Learning}

Prompt learning \citep{zhou2022learning,zhou2022conditional,li2025advancing,yang2024mma,khattak2023maple,yao2023visual, du2024ipo,10171397,10814093,danish2025comprehensive, zheng2025group} has become a widely adopted strategy for adapting large pre-trained VLMs to downstream tasks with limited labeled data. By optimizing only a small set of prompt parameters---typically continuous embeddings---while keeping the backbone frozen, prompt learning achieves effective task adaptation with relatively low computational cost. Representative methods such as CoOp \citep{zhou2022learning} and CoCoOp \citep{zhou2022conditional} have demonstrated strong few-shot performance and improved transferability. However, because these prompts are learned entirely in a continuous embedding space, the resulting prompt vectors are often difficult to interpret and cannot be directly mapped to meaningful natural-language semantics.

To improve the interpretability of prompt learning, recent studies have introduced explicit semantic structures into prompt design \citep{li2025advancing,ding2024tree,ghosal2024intcoop,du2024ipo,kim2024aapl}. TAP \citep{ding2024tree}, for example, constructs an LLM-generated tree of attributes and combines it with learnable expert tokens and vision-conditional pooling to achieve attribute-level alignment. IntCoOp \citep{ghosal2024intcoop} extends CoOp by incorporating a graph-based prompt interaction mechanism based on class relations, thereby enhancing both interpretability and classification performance. ATPrompt \citep{li2025advancing} anchors prompts with fixed universal attribute tokens and employs differentiable search for token selection, improving generalization to unseen categories. IPO \citep{du2024ipo} uses LMM-generated image descriptions and an LLM to iteratively produce human-readable, dataset-specific prompts without gradient-based learning, improving both interpretability and transferability.

Despite these advances, existing interpretable prompt learning methods often depend on external knowledge sources, such as LLMs, LMMs, or predefined attribute inventories. Such reliance increases annotation or inference cost and may reduce scalability in practical deployment. In some cases, these methods also suffer from instability when handling long textual inputs or complex semantic descriptions \citep{du2024ipo}. In contrast, IPL selects a small set of discrete semantic tokens from a filtered candidate vocabulary and incorporates them into prompts as fixed semantic anchors interleaved with learnable context vectors. This design introduces explicit semantic grounding without relying on costly external generation or elaborate attribute engineering, while maintaining a favorable balance among interpretability, adaptability, and efficiency.

\begin{figure}
    \centering
    \includegraphics[width=\textwidth]{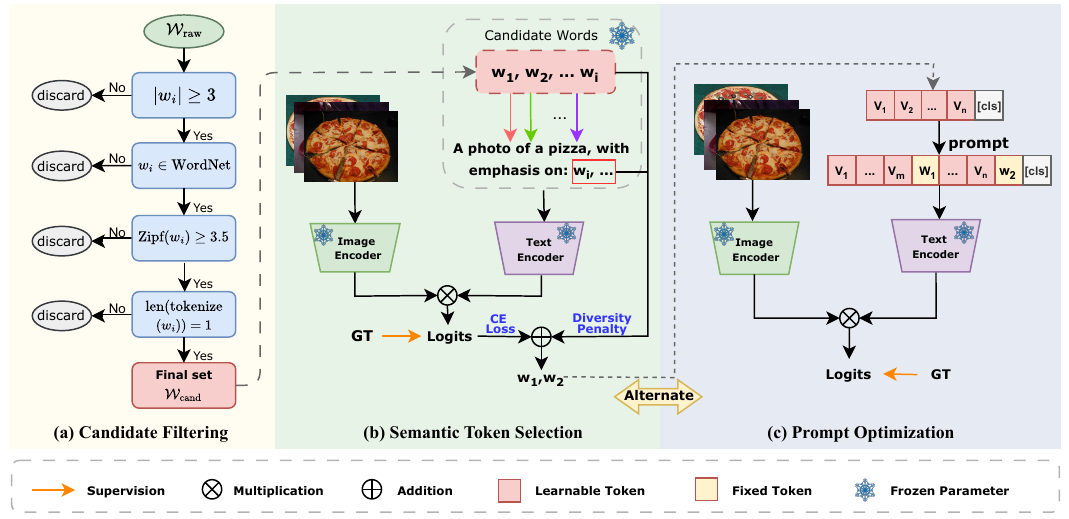}
    \caption{Pipeline of our method, divided into three main stages. (a) We begin by filtering the raw word set through a series of criteria to construct a refined candidate pool. (b) From this pool, we perform greedy token selection to identify semantically relevant tokens, which are inserted into prompt and serve as interpretable tokens that guide the prompt learning. (c) We alternate between semantic token selection and continuous prompt optimization to integrate the two process adaptively.}
    \label{fig:pipeline}
\end{figure}

\subsection{Submodular Optimization}
Submodular optimization \citep{liu2020submodular,zhu2024submodular,dughmi2009submodular} addresses discrete problems defined by the diminishing returns property: the marginal gain of adding an element to a set decreases as the set grows. Formally, a set function \( F: 2^V \rightarrow \mathbb{R} \) is submodular if, for all $A \subseteq B \subseteq V$ and any $v \in V \setminus B$, it holds that:
\begin{equation}
F(A \cup \{v\}) - F(A) \geq F(B \cup \{v\}) - F(B).
\end{equation}

A key advantage of submodular functions is that greedy algorithms can provide efficient approximate solutions. For monotone, non-negative functions under a cardinality constraint, the greedy approach guarantees a \((1\!-\!1/e)\)-approximation to the optimal solution \citep{nemhauser1978analysis,krause2014submodular}. Submodular optimization has been widely applied in areas such as active learning \citep{kothawade2021similar,golovin2011adaptive}, data summarization \citep{tschiatschek2014learning,mitrovic2018data,schreiber2020apricot}, as well as exemplar selection for few-shot learning \citep{killamsetty2021glister}.

In practice, strict submodularity is often difficult to achieve, but approximately submodular functions retain many of their desirable properties \citep{chierichetti2022additive,ajayi2019approximate,das2018approximate}. A function \( f: 2^V \rightarrow \mathbb{R} \) is said to be $\epsilon$-approximately submodular if there exists a submodular function $F$ such that:
\begin{equation}
\label{aproximate}
(1 - \epsilon) F(S) \leq f(S) \leq (1 + \epsilon) F(S),
\end{equation}
for all subsets \( S \subseteq V \), where $\epsilon$ denotes the approximation error, which represents the deviation from strict submodularity. Such functions exhibit diminishing marginal gains in a relaxed sense, allowing greedy algorithms to achieve near-optimal solutions with provable guarantees. Recent studies \citep{horel2016maximization,bian2017guarantees} show that greedy methods remain effective for approximate or near-submodular functions, often yielding strong empirical performance.

Motivated by these observations, we adopt an approximately submodular objective to guide semantic token selection. This formulation provides a practical basis for greedy search, allowing us to balance efficiency and solution quality while capturing both discriminative utility and semantic diversity.

\section{Problem Setup}

\subsection{The Prediction Function in CLIP}
CLIP \citep{radford2021learning} consists of a visual encoder and a textual encoder. The visual encoder maps an image to a high-dimensional embedding, while the textual encoder converts a prompt into a textual embedding. Both embeddings are aligned in a shared space to enable direct image--text comparison.

For $N$ categories, CLIP generates textual features by inserting each class label (e.g., ``pizza'') into a handcrafted template $P_h$ (``a photo of a [CLS]'') and encoding it with $E_\text{text}$, yielding $\mathbf{y}=E_\text{text}(P_h)$. Meanwhile, for an image $\mathbf{x}$, the visual encoder $E_\text{img}$ produces $\mathbf{u}=E_\text{img}(x)$. Classification for each class $c$ is then performed by computing cosine similarities between $\mathbf{u}$ and all $\mathbf{y}_c$, followed by a softmax:
\begin{equation}
\label{eq:clip}
p(c|x) = \frac{\exp(\text{sim}(\mathbf{u}, \mathbf{y}_c) / \tau)}{\sum_{i=1}^{N} \exp(\text{sim}(\mathbf{u}, \mathbf{y}_i) / \tau)},
\end{equation}
where $\text{sim}(\cdot, \cdot)$ denotes cosine similarity, and $\tau$ is a learnable temperature parameter controlling the sharpness of the distribution.

\subsection{Prompt Structure}
In prompt learning, prompts are designed to adapt pretrained models to downstream tasks, typically through learnable context tokens. In CoOp \citep{zhou2022learning}, the original prompt $P$ consists of $n$ learnable soft tokens followed by the class name $\text{[CLS]}$:
\begin{equation}
P = [V_1] [V_2] \dots [V_n] \, [\text{CLS}],
\end{equation}
where $\{V_i \in \mathbb{R}^d\}_{i=1}^n$ are learnable prompt embeddings, and $d$ is the embedding dimension. These vectors are optimized end-to-end with the contrastive objective, enabling automatic adaptation of prompt context to downstream datasets.

In our method, let $W_{\text{cand}}$ denote the candidate set filtered from the raw corpus $W_{\text{raw}}$. A subset of tokens is selected from $W_{\text{cand}}$ and inserted into the prompt $P$, yielding a modified prompt $\tilde{P}$. These selected tokens remain fixed during optimization. For example, when two semantic tokens are chosen, the input embedding sequence is interleaved as:
\begin{equation}
\label{eq:prompt_structure}
\tilde{P} = [V_1]\cdots[V_m][w_1][V_{m+1}]\cdots[V_{2m}][w_2]\cdots[V_n][\text{CLS}],
\end{equation}
where each semantic token $w_i$ follows a group of $m$ learnable tokens \citep{li2025advancing}. This alternating structure preserves both task adaptability and semantic stability during optimization, thereby balancing adaptability and interpretability throughout prompt learning.

\section{The Proposed Method}

We propose a framework that alternates between discrete semantic token selection and continuous prompt optimization. Semantic tokens are selected through a greedy algorithm guided by an approximate submodular objective and then inserted into the prompt as discrete semantic anchors, while the remaining soft context tokens are continuously optimized. This alternating process balances interpretability and adaptability. An overview of the pipeline is shown in Figure~\ref{fig:pipeline}.

\subsection{Candidate Filtering}
To enable effective semantic selection, we construct a high-quality candidate pool of interpretable tokens from the NLTK corpus \citep{bird2009natural}. This step ensures that the selected tokens carry human-understandable meanings while remaining compatible with both linguistic and tokenization constraints. Specifically, we apply four filtering criteria to derive the candidate set $W_{\text{cand}}$ from the raw corpus $W_{\text{raw}}$.

First, we retain only words that are purely alphabetic and satisfy $|w_i| \geq 3$, excluding symbols, single letters, and other non-standard textual artifacts that are unlikely to provide semantically meaningful information. Second, to ensure semantic validity, each candidate word must satisfy $w_i \in \text{WordNet}$, where WordNet \citep{miller1995wordnet} serves as an external lexical database distinct from the NLTK word list and helps filter out non-semantic entries such as personal names or place names. This constraint ensures that all selected tokens correspond to concrete and interpretable concepts. Third, we impose a frequency threshold requiring $\text{Zipf}(w_i) \geq 3.5$, where the Zipf \citep{van2014subtlex} score reflects how commonly a word is used in natural language. This removes rare or obscure terms that may hinder generalization. Finally, to maintain consistency with the tokenization behavior of the CLIP text encoder, we require that $\text{len}(\text{tokenize}(w_i)) = 1$, ensuring that each word maps to exactly one token under the CLIP tokenizer. This avoids semantic fragmentation during tokenization and preserves the integrity of each selected concept in the embedding process. The first three criteria correspond to linguistic constraints, while the last one concerns tokenization consistency.

Formally, the final filtered set is:
\begin{equation}
W_{\text{cand}} = \left\{ w_i \in W_{\text{raw}} \;\middle|\;
\begin{array}{l}
|w_i| \geq 3, \\
w_i \in \text{WordNet}, \\
\text{Zipf}(w_i) \geq 3.5, \\
\text{len}(\text{tokenize}(w_i)) = 1
\end{array}
\right\}
\end{equation}

The resulting candidate pool contains 8,883 tokens, forming a compact yet diverse set of interpretable words. This size provides sufficient semantic coverage while keeping computational cost manageable, making it well suited for semantic token selection. Based on this curated vocabulary, the subsequent optimization can be performed without relying on external models, enabling an efficient and self-contained pipeline.

\begin{figure}
    \centering
    \includegraphics[width=0.6\textwidth]{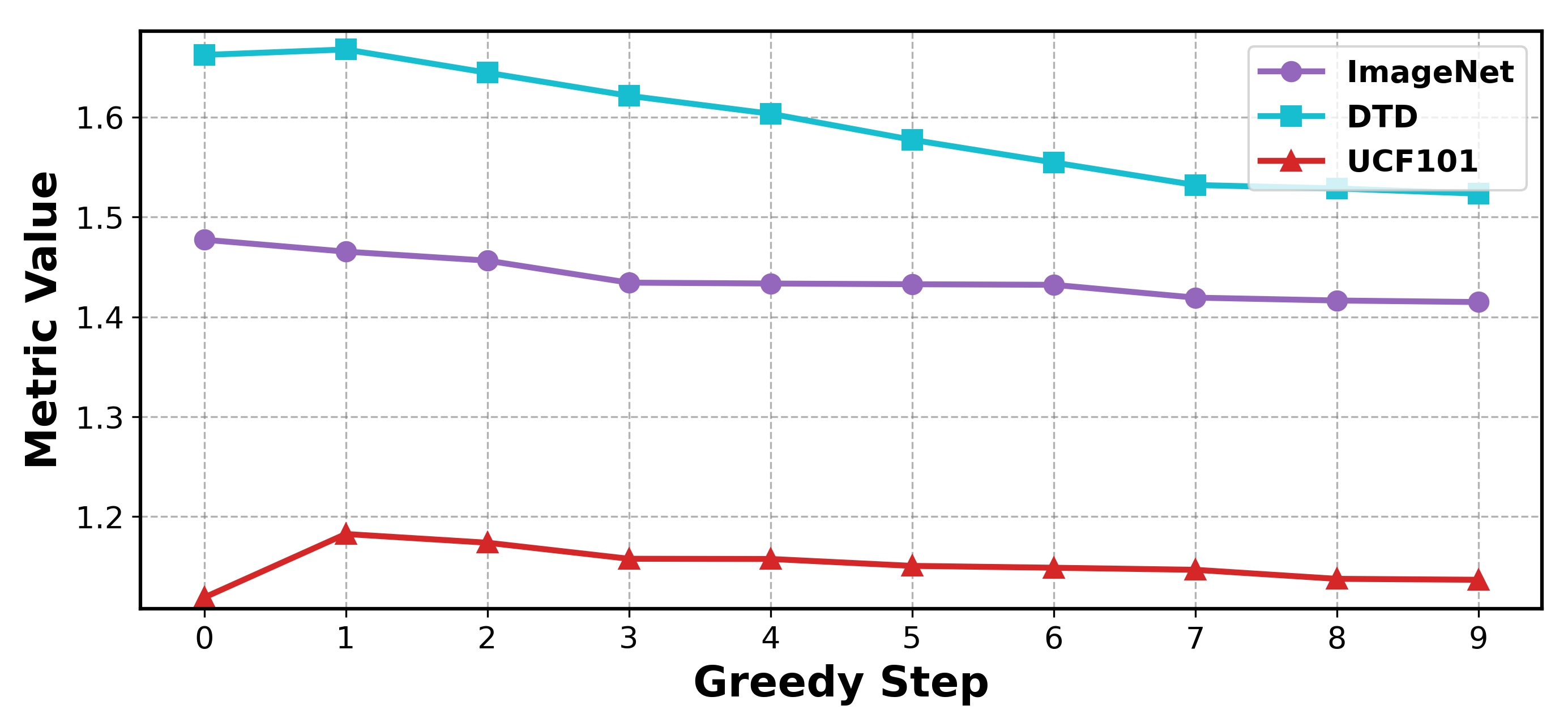}
    \caption{Empirical diminishing returns of marginal gain in token selection, showing the decreasing trend as tokens are added.}
    \label{fig:marginal}
\end{figure}

\subsection{Semantic Token Selection}\label{sec:selection}

With the candidate pool prepared, we next introduce the semantic token selection process. Our goal is to construct a token subset that is both class-relevant and semantically diverse. To this end, we define a set-level objective that combines a utility term and a redundancy penalty:
\begin{equation}
\label{eq:G_def}
G(W_{\text{sel}})=\mathrm{Utility}(W_{\text{sel}})-\lambda\,\mathrm{Redundancy}(W_{\text{sel}}),
\end{equation}
where \(W_{\text{sel}}\) denotes the currently selected token set and \(\lambda\) controls the trade-off between discriminability and diversity.

For the utility term, we measure how much the selected tokens improve image--text alignment under a CLIP-based prompt. Specifically, let
\[
P_{\text{sel}}=\text{``a photo of a [CLS], with emphasis on: [\dots]''},
\]
where \([\dots]\) is replaced by the tokens in \(W_{\text{sel}}\). We define
\begin{equation}
\label{eq:utility_def}
\mathrm{Utility}(W_{\text{sel}})
=
\mathcal{L}_{\mathrm{ce}}(\emptyset)-\mathcal{L}_{\mathrm{ce}}(W_{\text{sel}}),
\end{equation}
where \(\mathcal{L}_{\mathrm{ce}}(W_{\text{sel}})\) is the CLIP-based classification loss under prompt \(P_{\text{sel}}\), and \(\mathcal{L}_{\mathrm{ce}}(\emptyset)\) denotes the corresponding loss without any inserted semantic tokens. A larger utility indicates a larger reduction in classification loss brought by the selected tokens.

The classification loss is defined as
\begin{equation}
\label{eq:ce_loss}
\mathcal{L}_{\mathrm{ce}}(W_{\text{sel}})
=
-\log
\frac{\exp(\mathrm{sim}(\mathbf{u}_i,\mathbf{y}_c(W_{\text{sel}}))/\tau)}
{\sum_{j=1}^{N}\exp(\mathrm{sim}(\mathbf{u}_i,\mathbf{y}_j(W_{\text{sel}}))/\tau)},
\end{equation}
where \(\mathbf{u}_i\) denotes the CLIP image feature of the \(i\)-th sample, \(\mathbf{y}_j(W_{\text{sel}})\) denotes the text embedding of the prompt instantiated with token set \(W_{\text{sel}}\) for class \(j\), and \(\tau\) is the temperature parameter.

To encourage diversity, we define the redundancy term as the sum of pairwise similarities among selected tokens:
\begin{equation}
\label{eq:red_def}
\mathrm{Redundancy}(W_{\text{sel}})
=
\sum_{\substack{w_i, w_j \subseteq W_{\text{sel}} \\ i<j}}
\mathrm{sim}(\mathbf{e}_i,\mathbf{e}_j),
\end{equation}
where \(\mathbf{e}_i\) and \(\mathbf{e}_j\) denote the embeddings of selected tokens \(w_i\) and \(w_j\). A smaller redundancy indicates less semantic overlap within the selected set.

Our formulation is motivated by approximately diminishing marginal behavior in practice: as more informative tokens are added, the additional reduction in classification loss typically becomes smaller, while semantically similar tokens incur increasing redundancy. As shown in Figure~\ref{fig:marginal}, the marginal gain generally exhibits a decreasing trend as the greedy selection proceeds. Therefore, although \(G(W_{\text{sel}})\) does not strictly satisfy submodularity, it can be viewed as approximately submodular in practice and is thus well suited to greedy optimization.

At step \(l\), given the current selected set \(W_{\text{sel}}^{(l-1)}\), we greedily select the token that maximizes the marginal objective gain:
\begin{equation}
\label{eq:greedy}
w^*
=
\arg\max_{w \in W_{\text{cand}} \setminus W_{\text{sel}}^{(l-1)}}
\Big[
\Delta \mathrm{Utility}(w \mid W_{\text{sel}}^{(l-1)})
-
\lambda\,\Delta \mathrm{Redundancy}(w \mid W_{\text{sel}}^{(l-1)})
\Big].
\end{equation}
Here, the marginal utility gain is defined as
\begin{equation}
\label{eq:delta_utility}
\Delta \mathrm{Utility}(w \mid W_{\text{sel}}^{(l-1)})
=
\mathcal{L}_{\mathrm{ce}}(W_{\text{sel}}^{(l-1)})
-
\mathcal{L}_{\mathrm{ce}}(W_{\text{sel}}^{(l-1)} \cup \{w\}),
\end{equation}
which measures the additional reduction in classification loss after adding \(w\). The marginal redundancy increase is defined as
\begin{equation}
\label{eq:delta_redundancy}
\Delta \mathrm{Redundancy}(w \mid W_{\text{sel}}^{(l-1)})
=
\sum_{w_i \subseteq W_{\text{sel}}^{(l-1)}}
\mathrm{sim}(\mathbf{e}_w,\mathbf{e}_i),
\end{equation}
which measures the semantic overlap introduced by \(w\) with respect to the current selected set.

That is, the selected token should maximally reduce the classification loss while introducing as little semantic redundancy as possible. After each step, $w^*$ is removed from the candidate pool to avoid duplicate selection.

\subsection{Prompt Learning with Alternating Optimization}

To integrate discrete token selection with continuous prompt tuning, we adopt an alternating optimization strategy that balances interpretability and adaptability. Specifically, discrete token selection and continuous prompt learning are performed iteratively until $k$ tokens have been selected, ensuring both semantic clarity and flexibility in prompt representation.

In the continuous stage, the soft tokens in the prompt are optimized via gradient-based updates. During each iteration, a batch of image--text pairs is passed through the model, and prediction scores are computed in the shared embedding space. Formally, for a single image, the training objective follows the same contrastive loss as before:
\begin{equation}
\mathcal{L}
=
-\log
\frac{\exp(\cos(\mathbf{u}_i, \mathbf{y}_c) / \tau)}
{\sum_{j=1}^{N} \exp(\cos(\mathbf{u}_i, \mathbf{y}_j) / \tau)},
\end{equation}
where $\mathbf{y}_j$ and $\mathbf{y}_c$ denote the textual embeddings generated from the current prompt $\tilde{P}$ in Eq.~(\ref{eq:prompt_structure}). The gradients of this loss are then backpropagated to update the learnable soft prompt tokens.

\begin{algorithm}
\caption{Semantic Token Selection and Prompt Optimization}
\label{alg:selection_train}
\begin{algorithmic}[1]
\REQUIRE number of selected tokens $k$, learning rate $\alpha$, candidate pool $W_{\text{cand}}$, total epochs $T$, alternating interval $t$
\STATE \textbf{Initialize} selected token set $W_{\text{sel}} \gets \emptyset$, prompt parameters $\tilde{P} \gets$ random initialization
\FOR{$q = 1$ \textbf{to} $k$}
    \STATE Select $w^*$ according to Eq.~(\ref{eq:greedy})
    \STATE Update $W_{\text{sel}} \gets W_{\text{sel}} \cup \{w^*\}$ and $W_{\text{cand}} \gets W_{\text{cand}} \setminus \{w^*\}$
    \STATE Update the prompt with the current selected token set $W_{\text{sel}}$
    \FOR{$d = 1$ \textbf{to} $t$}
        \STATE Optimize $\tilde{P} \gets \tilde{P} - \alpha \nabla \mathcal{L}$
    \ENDFOR
\ENDFOR
\FOR{$r = 1$ \textbf{to} $T - kt$}
    \STATE Optimize $\tilde{P} \gets \tilde{P} - \alpha \nabla \mathcal{L}$
\ENDFOR
\RETURN $\tilde{P}, W_{\text{sel}}$
\end{algorithmic}
\end{algorithm}

In the discrete stage, one semantic token is selected from the candidate pool in each iteration following Eq.~(\ref{eq:greedy}). This selection step allows us to gradually introduce relevant semantic tokens into the prompt.

We then integrate the two stages into a unified alternating optimization procedure, which is conducted for a total of $T$ epochs. Specifically, after selecting one token, it is inserted into the prompt, resulting in an updated prompt $\tilde{P}$. The model is then optimized for $t$ epochs via continuous optimization, allowing it to adapt to the newly introduced token. This process repeats until $k$ semantic tokens have been incorporated. Once all semantic tokens are selected, the remaining $T - kt$ epochs are devoted to refining the learnable context vectors while keeping the selected tokens fixed. This alternating strategy mitigates early selection bias, improves the consistency between token semantics and downstream objectives, and ultimately enhances generalization.

The pseudocode for our approach is summarized in Algorithm~\ref{alg:selection_train}, which illustrates the process of semantic token selection and optimization. The algorithm operates in three phases:

\begin{itemize}
    \item Semantic Token Selection (lines~2--5): At each iteration, the algorithm selects a new semantic token from $W_{\text{cand}}$ according to Eq.~(\ref{eq:greedy}).

    \item Alternating Optimization (lines~6--8): After each token is selected, the prompt is updated and the learnable soft prompt tokens are optimized for $t$ epochs.

    \item Final Refinement (lines~10--12): Once all semantic tokens have been selected, the learnable soft prompt tokens are further optimized for the remaining $T - kt$ epochs.

\end{itemize}

\section{Experiments and Analysis}

\subsection{Setup}

\subsubsection{Dataset}
To evaluate both base-to-novel transfer and cross-dataset generalization, we conduct experiments on 11 widely used image classification benchmarks covering diverse recognition scenarios. The evaluation includes generic object recognition (ImageNet \citep{deng2009imagenet}, Caltech101 \citep{fei2004learning}); fine-grained categorization across animals, vehicles, plants, and food (OxfordPets \citep{parkhi2012cats}, StanfordCars \citep{krause20133d}, Flowers102 \citep{nilsback2008automated}, Food101 \citep{bossard2014food}, FGVCAircraft \citep{maji2013fine}); scene understanding (SUN397 \citep{xiao2010sun}); action recognition (UCF101 \citep{soomro2012ucf101}); texture classification (DTD \citep{cimpoi2014describing}); and remote sensing imagery (EuroSAT \citep{helber2019eurosat}).

To further assess robustness under distribution shifts, we adopt ImageNet as the source dataset and evaluate on four ImageNet variants: ImageNet-V2 \citep{recht2019imagenet}, ImageNet-Sketch \citep{wang2019learning}, ImageNet-A \citep{hendrycks2021natural}, and ImageNet-R \citep{hendrycks2021many}.

\begin{table}[t]
\caption{Base-to-Novel Generalization Results}
\centering
\label{tab:btone_general}

\footnotesize
\resizebox{\textwidth}{!}{%
\begin{tabular}{lcccccccccccc}
\toprule
\multirow{2}{*}{Methods}
    & \multicolumn{3}{c}{Average}
    & \multicolumn{3}{c}{ImageNet}
    & \multicolumn{3}{c}{Caltech101}
    & \multicolumn{3}{c}{OxfordPets} \\
\cmidrule(lr){2-4} \cmidrule(lr){5-7} \cmidrule(lr){8-10} \cmidrule(lr){11-13}
    & Base & Novel & HM
    & Base & Novel & HM
    & Base & Novel & HM
    & Base & Novel & HM \\
\midrule
CoOp       & 82.69 & 63.22 & 71.66  & 76.47 & 67.88 & 71.92  & 98.00 & 89.81 & 93.73  & 93.67 & 95.29 & 94.47 \\
\textbf{+IPL}      & 82.49 & 70.00 & 75.73 \scriptsize{\textcolor{cyan}{(+4.07)}} & 76.38 & 66.07 & 70.85 \scriptsize{\textcolor{gray}{(-1.07)}} & 98.00 & 92.53 & 95.19 \scriptsize{\textcolor{cyan}{(+1.82)}} & 95.48 & 96.50 & 95.99 \scriptsize{\textcolor{cyan}{(+1.52)}} \\
CoCoOp     & 80.47 & 71.69 & 75.83  & 75.98 & 70.43 & 73.10  & 97.96 & 93.81 & 95.84  & 95.20 & 97.69 & 96.43 \\
\textbf{+IPL}    & 81.52 & 73.00 & 77.02 \scriptsize{\textcolor{cyan}{(+1.19)}} & 76.34 & 70.12 & 73.10 \scriptsize{\textcolor{cyan}{(+0.00)}} & 97.78 & 94.79 & 96.26 \scriptsize{\textcolor{cyan}{(+0.42)}} & 95.85 & 97.39 & 96.61 \scriptsize{\textcolor{cyan}{(+0.18)}} \\
KgCoOp     & 80.73 & 73.60 & 77.00  & 75.83 & 69.96 & 72.78  & 97.72 & 94.39 & 96.03  & 94.65 & 97.76 & 96.18 \\
\textbf{+IPL}   & 80.37 & 74.43 & 77.29 \scriptsize{\textcolor{cyan}{(+0.29)}} & 76.17 & 70.12 & 73.02 \scriptsize{\textcolor{cyan}{(+0.24)}} & 97.77 & 94.36 & 96.03 \scriptsize{\textcolor{cyan}{(+0.00)}} & 95.10 & 97.96 & 96.51 \scriptsize{\textcolor{cyan}{(+0.33)}} \\
MaPLe      & 82.28 & 75.14 & 78.55  & 76.66 & 70.54 & 73.47  & 97.74 & 94.36 & 96.02  & 95.43 & 97.76 & 96.58 \\
\textbf{+IPL}     & 83.51 & 74.85 & 78.94 \scriptsize{\textcolor{cyan}{(+0.39)}} & 76.90 & 70.36 & 73.48 \scriptsize{\textcolor{cyan}{(+0.01)}} & 98.19 & 94.18 & 96.14 \scriptsize{\textcolor{cyan}{(+0.12)}} & 95.59 & 97.67 & 96.62 \scriptsize{\textcolor{cyan}{(+0.04)}} \\
PromptSRC  & 84.26 & 76.10 & 79.97 & 77.60 & 70.73 & 74.01  & 98.10 & 94.03 & 96.02  & 95.33 & 97.30 & 96.30 \\
\textbf{+IPL} & 84.46 & 76.58 & 80.33 \scriptsize{\textcolor{cyan}{(+0.36)}} & 77.69 & 70.66 & 74.01 \scriptsize{\textcolor{cyan}{(+0.00)}} & 98.13 & 94.29 & 96.17 \scriptsize{\textcolor{cyan}{(+0.15)}} & 95.85 & 97.61 & 96.72 \scriptsize{\textcolor{cyan}{(+0.42)}} \\
\bottomrule
\end{tabular}%
}

\vspace{1ex}

\footnotesize
\resizebox{\textwidth}{!}{%
\begin{tabular}{lcccccccccccc}
\toprule
\multirow{2}{*}{Methods}
    & \multicolumn{3}{c}{StanfordCars}
    & \multicolumn{3}{c}{Flowers102}
    & \multicolumn{3}{c}{Food101}
    & \multicolumn{3}{c}{FGVCAircraft} \\
\cmidrule(lr){2-4} \cmidrule(lr){5-7} \cmidrule(lr){8-10} \cmidrule(lr){11-13}
    & Base & Novel & HM
    & Base & Novel & HM
    & Base & Novel & HM
    & Base & Novel & HM \\
\midrule
CoOp       & 78.12 & 60.40 & 68.13  & 97.60 & 59.67 & 74.06  & 88.33 & 82.26 & 85.19  & 40.44 & 22.30 & 28.75 \\
\textbf{+IPL}     & 77.38 & 68.61 & 72.73 \scriptsize{\textcolor{cyan}{(+4.60)}} & 97.18 & 70.47 & 81.70 \scriptsize{\textcolor{cyan}{(+7.64)}} & 89.68 & 88.93 & 89.30 \scriptsize{\textcolor{cyan}{(+4.11)}} & 39.10 & 30.43 & 34.22 \scriptsize{\textcolor{cyan}{(+5.47)}} \\
CoCoOp     & 70.49 & 73.59 & 72.01  & 94.87 & 71.75 & 81.71  & 90.70 & 91.29 & 90.99  & 33.41 & 23.71 & 27.74 \\
\textbf{+IPL}    & 73.14 & 72.47 & 72.80 \scriptsize{\textcolor{cyan}{(+0.79)}} & 96.26 & 70.89 & 82.81 \scriptsize{\textcolor{cyan}{(+1.10)}} & 90.41 & 91.38 & 90.89 \scriptsize{\textcolor{gray}{(-0.10)}} & 36.61 & 32.33 & 34.32 \scriptsize{\textcolor{cyan}{(+6.58)}} \\
KgCoOp     & 71.76 & 75.04 & 73.36  & 95.00 & 74.73 & 83.65  & 90.50 & 91.70 & 91.09  & 36.21 & 33.55 & 34.83 \\
\textbf{+IPL}    & 72.17 & 74.83 & 73.47 \scriptsize{\textcolor{cyan}{(+0.11)}} & 93.77 & 75.68 & 83.76 \scriptsize{\textcolor{cyan}{(+0.11)}} & 90.52 & 91.72 & 91.11 \scriptsize{\textcolor{cyan}{(+0.02)}} & 34.74 & 35.09 & 34.91 \scriptsize{\textcolor{cyan}{(+0.08)}} \\
MaPLe      & 72.94 & 74.00 & 73.47  & 95.92 & 72.46 & 82.56  & 90.71 & 92.05 & 91.38  & 37.44 & 35.61 & 36.50 \\
\textbf{+IPL}    & 76.29 & 73.35 & 74.79 \scriptsize{\textcolor{cyan}{(+1.32)}} & 96.24 & 73.10 & 83.66 \scriptsize{\textcolor{cyan}{(+1.10)}} & 90.45 & 91.67 & 91.05 \scriptsize{\textcolor{gray}{(-0.33)}} & 40.66 & 35.71 & 37.98 \scriptsize{\textcolor{cyan}{(+1.48)}} \\
PromptSRC  & 78.27 & 74.97 & 73.47  & 98.07 & 76.50 & 85.95  & 90.67 & 91.53 & 91.10  & 42.73 & 37.87 & 40.15 \\
\textbf{+IPL}  & 78.96 & 75.11 & 76.99 \scriptsize{\textcolor{cyan}{(+3.52)}} & 97.88 & 77.97 & 86.79 \scriptsize{\textcolor{cyan}{(+0.84)}} & 90.75 & 91.62 & 91.19 \scriptsize{\textcolor{cyan}{(+0.09)}} & 42.64 & 38.29 & 40.33 \scriptsize{\textcolor{cyan}{(+0.18)}} \\
\bottomrule
\end{tabular}%
}

\vspace{1ex}

\footnotesize
\resizebox{\textwidth}{!}{%
\begin{tabular}{lcccccccccccc}
\toprule
\multirow{2}{*}{Methods}
    & \multicolumn{3}{c}{SUN397}
    & \multicolumn{3}{c}{DTD}
    & \multicolumn{3}{c}{EuroSAT}
    & \multicolumn{3}{c}{UCF101} \\
\cmidrule(lr){2-4} \cmidrule(lr){5-7} \cmidrule(lr){8-10} \cmidrule(lr){11-13}
    & Base & Novel & HM
    & Base & Novel & HM
    & Base & Novel & HM
    & Base & Novel & HM \\
\midrule
CoOp       & 80.60 & 65.89 & 72.51  & 79.44 & 41.18 & 54.24  & 92.19 & 54.74 & 68.69  & 84.69 & 56.05 & 67.46 \\
\textbf{+IPL}      & 81.16 & 70.40 & 75.40 \scriptsize{\textcolor{cyan}{(+2.89)}} & 80.17 & 52.30 & 63.30 \scriptsize{\textcolor{cyan}{(+9.06)}} & 89.22 & 67.73 & 77.10 \scriptsize{\textcolor{cyan}{(+8.41)}} & 83.68 & 66.03 & 73.82 \scriptsize{\textcolor{cyan}{(+6.36)}} \\
CoCoOp     & 79.74 & 76.68 & 78.27  & 77.01 & 56.00 & 64.85  & 87.49 & 60.04 & 71.21  & 82.33 & 73.45 & 77.64 \\
\textbf{+IPL}    & 80.79 & 76.64 & 78.66 \scriptsize{\textcolor{cyan}{(+0.39)}} & 79.32 & 55.79 & 65.51 \scriptsize{\textcolor{cyan}{(+0.66)}} & 86.66 & 66.09 & 74.59 \scriptsize{\textcolor{cyan}{(+3.38)}} & 83.57 & 75.07 & 79.09 \scriptsize{\textcolor{cyan}{(+1.45)}} \\
KgCoOp     & 80.29 & 76.53 & 78.36  & 77.55 & 54.99 & 64.35  & 85.64 & 64.34 & 73.48  & 82.89 & 76.67 & 79.65 \\
\textbf{+IPL}    & 78.52 & 78.12 & 78.32 \scriptsize{\textcolor{gray}{(-0.04)}} & 75.96 & 58.09 & 65.81 \scriptsize{\textcolor{cyan}{(+1.46)}} & 86.58 & 65.79 & 74.77 \scriptsize{\textcolor{cyan}{(+1.29)}} & 82.83 & 76.92 & 79.77 \scriptsize{\textcolor{cyan}{(+0.12)}} \\
MaPLe      & 80.82 & 78.70 & 79.75  & 80.36 & 59.18 & 68.16  & 94.07 & 73.23 & 82.35  & 83.00 & 78.66 & 80.77 \\
\textbf{+IPL}     & 81.81 & 77.83 & 79.77 \scriptsize{\textcolor{cyan}{(+0.02)}} & 81.33 & 57.51 & 67.38 \scriptsize{\textcolor{gray}{(-0.75)}} & 95.90 & 75.01 & 84.01 \scriptsize{\textcolor{cyan}{(+1.66)}} & 85.30 & 76.96 & 80.91 \scriptsize{\textcolor{cyan}{(+0.14)}} \\
PromptSRC  & 82.67 & 78.47 & 80.52  & 83.37 & 62.97 & 71.75  & 92.90 & 73.90 & 82.32  & 87.10 & 78.89 & 82.74 \\
\textbf{+IPL}  & 82.92 & 78.55 & 80.68 \scriptsize{\textcolor{cyan}{(+0.16)}} & 84.49 & 62.14 & 71.61 \scriptsize{\textcolor{gray}{(-0.14)}} & 92.73 & 77.13 & 84.18 \scriptsize{\textcolor{cyan}{(+1.86)}} & 87.02 & 79.01 & 82.82 \scriptsize{\textcolor{cyan}{(+0.08)}} \\
\bottomrule
\end{tabular}%
}
\end{table}

\subsubsection{Base-to-Novel Generalization}
Following existing protocols, we partition each dataset into base and novel classes, training exclusively on base classes and evaluating on both base and novel test sets. We adopt each baseline's original optimization schedule for fair comparison, reporting performance on both splits along with their harmonic mean (HM). This setting evaluates the model's ability to generalize from seen to unseen classes under distribution shifts.

\subsubsection{Cross-dataset Evaluation}
For cross-dataset evaluation, models are trained on ImageNet and evaluated on 10 unseen datasets from diverse domains without additional fine-tuning. We report accuracy on each dataset to assess the model's ability to transfer knowledge across different data distributions and visual domains.

\subsubsection{Domain Generalization}
To assess domain generalization, we evaluate ImageNet-trained models on four ImageNet variants: ImageNetV2, ImageNet-Sketch, ImageNet-A, and ImageNet-R. We report overall accuracy to measure model performance on out-of-distribution data, which is critical for real-world deployment.

\subsubsection{Implementation Details}
We use CLIP with a ViT-B/16 backbone for all experiments under the 16-shot setting, and report the average results over three runs. For CoOp and MaPLe, each fixed token is paired with $m=2$ learnable tokens, while other methods use $m=4$. Fixed tokens are initialized to zero and then replaced with selected tokens. All experiments are conducted on a single GPU using the original hyperparameter settings.

\renewcommand{\arraystretch}{0.85}
\begin{table}
\caption{Cross-Dataset Evaluation Results}
\centering
\label{tab:cross_dataset_full}
\resizebox{\textwidth}{!}{%
\begin{tabular}{lcccccccccccc}
\toprule
Method & \begin{tabular}[c]{@{}c@{}}Image\\Net\end{tabular}
& \begin{tabular}[c]{@{}c@{}}Caltech\\101\end{tabular}
& \begin{tabular}[c]{@{}c@{}}Oxford\\Pets\end{tabular}
& \begin{tabular}[c]{@{}c@{}}Stanford\\Cars\end{tabular}
& \begin{tabular}[c]{@{}c@{}}Flowers\\102\end{tabular}
& Food101
& \begin{tabular}[c]{@{}c@{}}FGVC\\Aircraft\end{tabular}
& SUN397
& DTD
& \begin{tabular}[c]{@{}c@{}}Euro\\SAT\end{tabular}
& UCF101
& Average \\
\midrule
CoOp   & 71.51 & 93.70 & 89.14 & 64.51 & 68.71 & 85.30 & 18.47 & 64.15 & 41.92 & 46.39 & 66.55 & 63.88 \\
+IPL   & 71.67 & 94.24 & 90.40 & 64.61 & 70.25 & 85.72 & 21.87 & 63.79 & 43.27 & 50.83 & 68.58 & \textbf{65.93} \scriptsize{\textcolor{cyan}{(+2.05)}} \\
\midrule
CoCoOp & 71.02 & 94.43 & 90.14 & 65.32 & 71.88 & 86.06 & 22.94 & 67.36 & 45.73 & 45.37 & 68.21 & 65.74 \\
+IPL   & 71.36 & 94.13 & 90.71 & 65.92 & 72.12 & 86.05 & 23.60 & 65.11 & 44.90 & 49.90 & 70.09 & \textbf{66.72} \scriptsize{\textcolor{cyan}{(+0.98)}} \\
\midrule
MaPLe  & 70.72 & 93.53 & 90.49 & 65.57 & 72.23 & 86.20 & 24.74 & 67.01 & 46.49 & 48.06 & 68.69 & 66.30 \\
+IPL   & 70.77 & 94.31 & 90.79 & 65.82 & 72.59 & 86.16 & 24.18 & 67.08 & 45.73 & 52.46 & 69.13 & \textbf{66.99} \scriptsize{\textcolor{cyan}{(+0.69)}} \\
\bottomrule
\end{tabular}%
}
\end{table}

\subsection{Main Result}

\subsubsection{Base-to-Novel Generalization}

We evaluate IPL on five representative baselines, including text-based methods—CoOp \citep{zhou2022learning}, CoCoOp \citep{zhou2022conditional}, KgCoOp \citep{yao2023visual}, PromptSrc \citep{khattak2023self}—and the multimodal approach MaPLe \citep{khattak2023maple}. Experiments are conducted on 11 datasets, reporting accuracy on base classes, novel classes, and their harmonic mean to assess generalization.

As shown in Table~\ref{tab:btone_general}, IPL consistently improves HM across all baselines. The most significant gains are observed on CoOp (+4.07\%) and CoCoOp (+1.19\%). Notable improvements are also achieved on several datasets, such as Flowers102 and EuroSAT, where the latter shows a +8.41\% increase on novel classes. These results indicate that IPL consistently enhances the balance between base and novel performance, demonstrating strong robustness and generalization across diverse scenarios.

\subsubsection{Cross-dataset Evaluation}
We further evaluate three representative baselines under a cross-dataset setting, where models are trained on ImageNet and tested on unseen datasets from different domains. This setting is particularly challenging due to substantial distribution shifts (e.g., from fine-grained object recognition to scene classification). Detailed results are reported in Table~\ref{tab:cross_dataset_full}.

Incorporating IPL leads to consistent performance gains, with average improvements of +2.05\%, +0.98\%, and +0.69\% for CoOp, CoCoOp, and MaPLe, respectively. Most datasets show clear improvements, indicating that IPL effectively enhances both model adaptability and zero-shot generalization ability.

\begin{table}
\caption{Domain Generalization Results}
\centering
\label{tab:domain_gen}

\resizebox{0.6\textwidth}{!}{%
\begin{tabular}{lcccccc}
\toprule
\multirow{2}{*}{Methods} & Source & \multicolumn{4}{c}{Target Dataset} & \multirow{2}{*}{Average} \\
\cmidrule(lr){3-6}
& ImageNet & -V2 & -S & -A & -R & \\
\midrule
CoOp   & 71.51 & 64.20 & 47.99 & 49.71 & 75.21 & 59.28 \\
+IPL   & 71.67 & 64.34 & 48.70 & 50.64 & 76.10 & 59.94 (\textcolor{cyan}{+0.66}) \\
\midrule
CoCoOp & 71.02 & 64.07 & 48.75 & 50.63 & 76.18 & 59.91 \\
+IPL   & 71.36 & 64.42 & 49.08 & 51.00 & 76.66 & 60.29 (\textcolor{cyan}{+0.38}) \\
\midrule
MaPLe  & 70.72 & 64.07 & 49.15 & 50.96 & 76.98 & 60.27 \\
+IPL   & 70.77 & 64.11 & 49.27 & 51.24 & 77.44 & 60.52 (\textcolor{cyan}{+0.25}) \\
\bottomrule
\end{tabular}%
}
\end{table}

\subsubsection{Domain Generalization}
To evaluate domain generalization, we integrate IPL into CoOp, CoCoOp, and MaPLe, train them on ImageNet, and test them on four out-of-domain datasets. As shown in Table~\ref{tab:domain_gen}, IPL consistently improves performance, yielding gains of +0.66\%, +0.38\%, and +0.25\% for the three baselines, respectively. These results demonstrate that IPL effectively enhances model robustness under domain shift and provides consistent benefits for domain generalization.

\subsection{Further Analysis}

\subsubsection{Semantic Alignment of Selected Tokens}

To illustrate the clarity and relevance of the identified tokens, Table~\ref{tab:token_only} presents the token selections across all datasets, together with human-annotated semantic relevance scores. The scores are provided by three annotators with undergraduate degrees, good English proficiency, and familiarity with the dataset contents. Each annotator rates the relevance between the selected tokens and the corresponding dataset on a 5-point scale, and the final score is obtained by averaging their ratings.

For \textit{OxfordPets}, the tokens convey physical traits and care-related concepts; in \textit{FGVCAircraft}, they relate to aircraft functionality and passenger experience; for \textit{Flower102}, they describe morphological features and aesthetic qualities; in \textit{UCF101}, they capture action dynamics and situational contexts; in \textit{EuroSAT}, the selected token highlights the dominance of color cues and surface patterns in satellite imagery; and in \textit{DTD}, the token ``heather'' reflects texture regularity and fine-grained material appearance. Overall, these examples show that our method can select tokens that are semantically rich and closely related to the conceptual, visual, and functional characteristics of each dataset, leading to more interpretable visual-textual associations.

However, for some datasets, the selected tokens are less informative, mainly due to high intra-class diversity or broad semantic coverage, as in \textit{SUN397} and \textit{ImageNet}. Many of these words have relatively low similarity scores, often around 0.5 (e.g., ``social'' with other candidate tokens), suggesting that the model tends to favor diversity when strong semantic cues are limited.

Compared with CoOp \citep{zhou2022learning}, which learns continuous context vectors without explicit semantic grounding, the nearest words retrieved from its embeddings (Table~\ref{tab:token_only}) often show limited visual relevance, including meaningless or dataset-specific fragments such as ``tosc,'' ``.\#,'' or ``nytimes.'' In contrast, IPL's discrete tokens are more consistent with domain semantics and receive higher human relevance scores in most cases, providing clearer and more interpretable associations between visual and textual concepts.

\begin{table}
\caption{Selected tokens for different datasets with human-annotated semantic relevance scores (1--5).}
\centering
\begin{tabular}{ccccc}
\toprule
Dataset & Selected Tokens & IPL Score & CoOp Tokens & CoOp Score \\
\midrule
\textbf{OxfordPets}   & furry, veterinary, adopt & 5 & tosc, enjoyment & 2 \\
\textbf{Flower102}    & flowering, tilted, stunning & 4 & - & - \\
\textbf{FGVCAircraft} & comfortable, touchdown & 4 & - & - \\
\textbf{UCF101}       & simulation, reaction, turning & 4 & facial, meteorologist & 2 \\
\textbf{EuroSAT}      & blue & 3 & - & - \\
\textbf{StanfordCars }         & departure & 3 & - & - \\
\textbf{Food101  }             & sweating, social & 2 & nytimes, lc & 1 \\
\textbf{DTD}          & heather, title, phone & 2 & honey, series & 1 \\
ImageNet              & visit, social & 1 & .\#, potd & 1 \\
Caltech101            & fascism, social, pile & 1 & - & - \\
SUN397                & excluding & 1 & - & - \\
\bottomrule
\end{tabular}
\label{tab:token_only}
\end{table}

\begin{table}
\caption{Effect of Removing Diversity Penalty on Model Performance}
\centering
\label{tab:diversity_penalty}
\resizebox{\textwidth}{!}{%
\begin{tabular}{lccccccccc}
\toprule
\multirow{2}{*}{\textbf{Dataset}} & \multicolumn{3}{c|}{\textbf{HM}} & \multicolumn{3}{c|}{\textbf{Base}} & \multicolumn{3}{c}{\textbf{New}} \\
 & \textbf{w/ Penalty} & \textbf{w/o Penalty} & \textbf{$\Delta$} & \textbf{w/ Penalty} & \textbf{w/o Penalty} & \textbf{$\Delta$} & \textbf{w/ Penalty} & \textbf{w/o Penalty} & \textbf{$\Delta$} \\
\midrule
ImageNet       & 70.85 & 69.33 & \textcolor{cyan}{+1.52} & 76.38 & 76.11 & \textcolor{cyan}{+0.27} & 66.07 & 63.67 & \textcolor{cyan}{+2.40} \\
Caltech101     & 95.19 & 94.28 & \textcolor{cyan}{+0.91} & 98.00 & 98.06 & \textcolor{gray}{-0.06} & 92.53 & 90.97 & \textcolor{cyan}{+1.56} \\
OxfordPets     & 97.18 & 94.52 & \textcolor{cyan}{+2.66} & 95.48 & 95.00 & \textcolor{cyan}{+0.48} & 96.50 & 94.18 & \textcolor{cyan}{+2.32} \\
StanfordCars   & 73.73 & 72.36 & \textcolor{cyan}{+1.37} & 77.38 & 78.20 & \textcolor{gray}{-0.82} & 68.61 & 67.35 & \textcolor{cyan}{+1.26} \\
Flowers102     & 81.70 & 80.53 & \textcolor{cyan}{+1.17} & 97.18 & 97.56 & \textcolor{gray}{-0.38} & 70.47 & 68.70 & \textcolor{cyan}{+1.77} \\
Food101        & 89.30 & 88.33 & \textcolor{cyan}{+0.97} & 89.68 & 88.84 & \textcolor{cyan}{+0.84} & 88.93 & 87.87 & \textcolor{cyan}{+1.06} \\
FGVCAircraft   & 34.22 & 32.75 & \textcolor{cyan}{+1.47} & 39.10 & 40.68 & \textcolor{gray}{-1.58} & 30.43 & 27.59 & \textcolor{cyan}{+2.84} \\
SUN397         & 75.40 & 72.57 & \textcolor{cyan}{+2.83} & 81.16 & 81.02 & \textcolor{cyan}{+0.14} & 70.40 & 65.72 & \textcolor{cyan}{+4.68} \\
DTD            & 63.30 & 62.03 & \textcolor{cyan}{+1.27} & 80.17 & 80.25 & \textcolor{gray}{-0.08} & 52.30 & 50.60 & \textcolor{cyan}{+1.70} \\
EuroSAT        & 77.10 & 69.87 & \textcolor{cyan}{+7.23} & 89.22 & 90.49 & \textcolor{gray}{-1.27} & 67.73 & 56.99 & \textcolor{cyan}{+10.74} \\
UCF101         & 73.82 & 71.36 & \textcolor{cyan}{+2.46} & 83.68 & 84.64 & \textcolor{gray}{-0.96} & 66.03 & 61.69 & \textcolor{cyan}{+4.34} \\
\midrule
\textbf{Average} & \textbf{75.62} & \textbf{73.45} & \textbf{+2.17} & \textbf{82.49} & \textbf{82.80} & \textbf{-0.31} & \textbf{70.00} & \textbf{66.85} & \textbf{+3.15} \\
\bottomrule
\end{tabular}%
}
\end{table}

\subsubsection{Diversity Penalty}
To evaluate the effect of diversity regularization, we remove the diversity penalty term from the token selection objective and conduct experiments on 11 benchmark datasets, using CoOp as the baseline under the same base-to-novel setting. As shown in Table~\ref{tab:diversity_penalty}, incorporating the diversity penalty consistently improves HM accuracy, suggesting its benefit for generalization. While base-class accuracy remains similar or slightly decreases, novel-class accuracy generally improves. Without this regularization, the selected tokens tend to overfit the semantics of base classes, limiting transfer to unseen categories. These results indicate that the diversity penalty helps promote a richer and more transferable set of semantic tokens.

Following the analysis in the previous section, we set $\lambda$ to 0.2 for datasets with clearer semantics (e.g., OxfordPets and Flowers102) and 0.1 for datasets with broader or less explicit semantics. However, when $\lambda$ is set too high (e.g., above 0.4), its effect becomes overly strong, especially on high-accuracy datasets such as Caltech101. In this case, the model tends to select less relevant tokens in pursuit of diversity, which reduces interpretability and weakens generalization. These findings highlight the importance of properly balancing the diversity penalty to achieve effective transfer across datasets.

\subsubsection{Token Quantity}

To further analyze the effect of token quantity, we vary the number of selected tokens $k$ and examine the resulting model behavior. While increasing $k$ enhances the semantic expressiveness of the prompt, we observe a continued reduction in the token selection loss $\mathcal{L}_\text{sel}$ beyond six tokens, indicating limited additional gains in semantic alignment. However, this reduction in $\mathcal{L}_\text{sel}$ does not lead to improved classification performance. Instead, inserting too many tokens tends to saturate the prompt space and introduce redundant or weakly relevant concepts.

To balance expressiveness and efficiency, we conduct a controlled evaluation across all datasets by varying $k \in \{1, \dots, 6\}$, and select the token quantity for each dataset based on training performance. Since the training and test sets are strictly separated, this selection does not involve any test-time information. This adaptive strategy ensures that the number of inserted semantic tokens is sufficient to capture dataset-specific representations while avoiding unnecessary prompt complexity that may hinder generalization.

\begin{figure}
    \centering
    \includegraphics[width=0.5\textwidth]{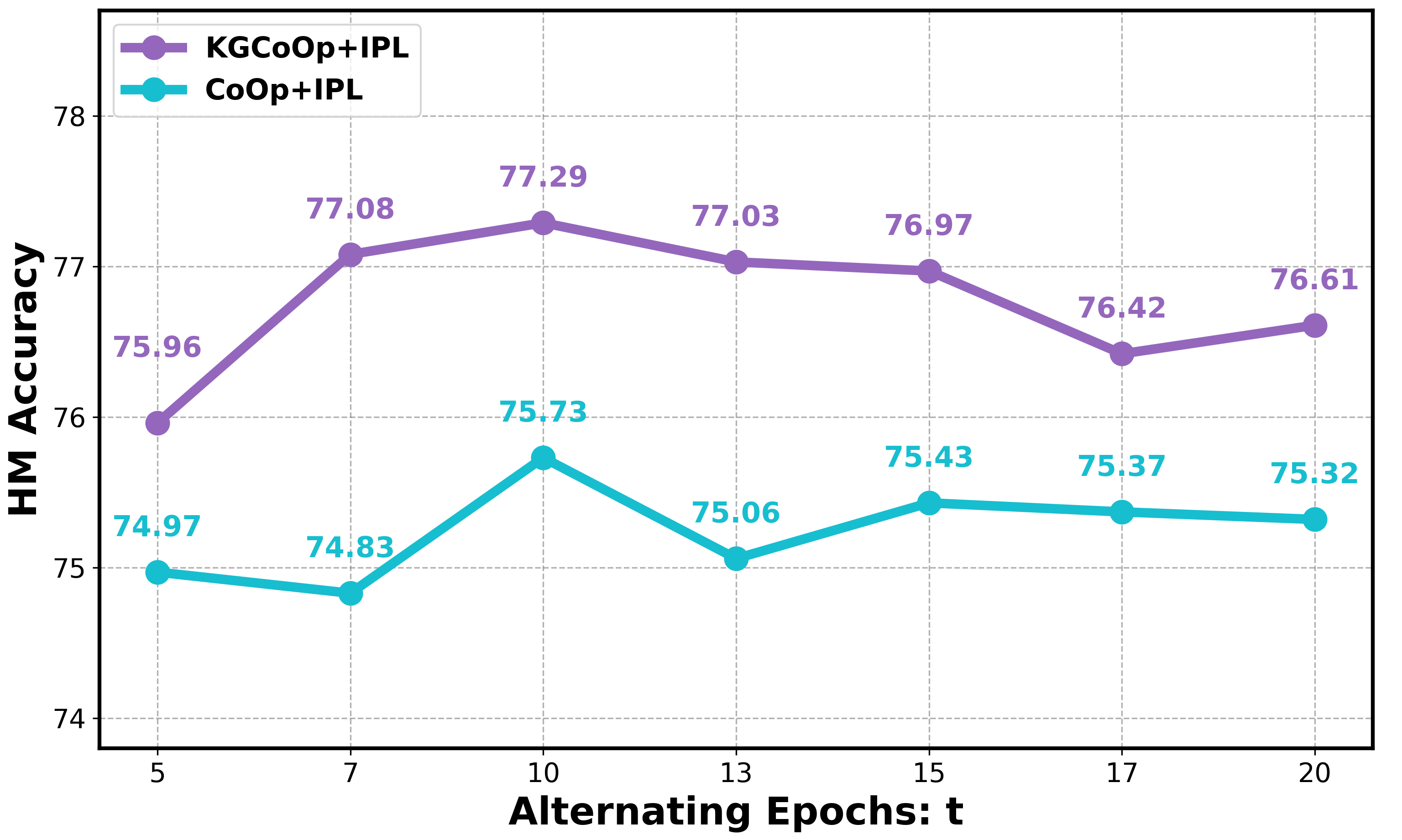}
    \caption{Effect of alternating interval $t$ on HM accuracy averaged over 11 datasets. The best performance is obtained at $t{=}10$, suggesting that balanced update scheduling benefits prompt learning stability and effectiveness.}
    \label{fig:epoch}
\end{figure}

\subsubsection{Alternating Epochs}

We study the effect of the alternating interval $t$ on representative baselines with relatively long training schedules, namely CoOp and KgCoOp, both trained for 100 epochs. We vary $t \in \{5,7,10,13,15,17,20\}$, where $t$ denotes the number of standard optimization epochs between two alternating updates.

As shown in Figure~\ref{fig:epoch}, $t=10$ consistently achieves the best or near-best HM accuracy on these baselines, indicating a favorable trade-off between base and novel performance. Both smaller and larger values of $t$ lead to slight performance degradation, suggesting that an appropriate update frequency is important.

For the other baselines, whose training schedules involve relatively few optimization epochs, the number of alternating updates is inherently limited. In such cases, varying $t$ does not provide a meaningful sensitivity analysis, as different choices result in only a small number of updates. We therefore set $t=1$ to apply semantic refinement as frequently as possible while preserving consistency with their original optimization protocols.

\begin{figure}
    \centering
    \includegraphics[width=0.6\textwidth]{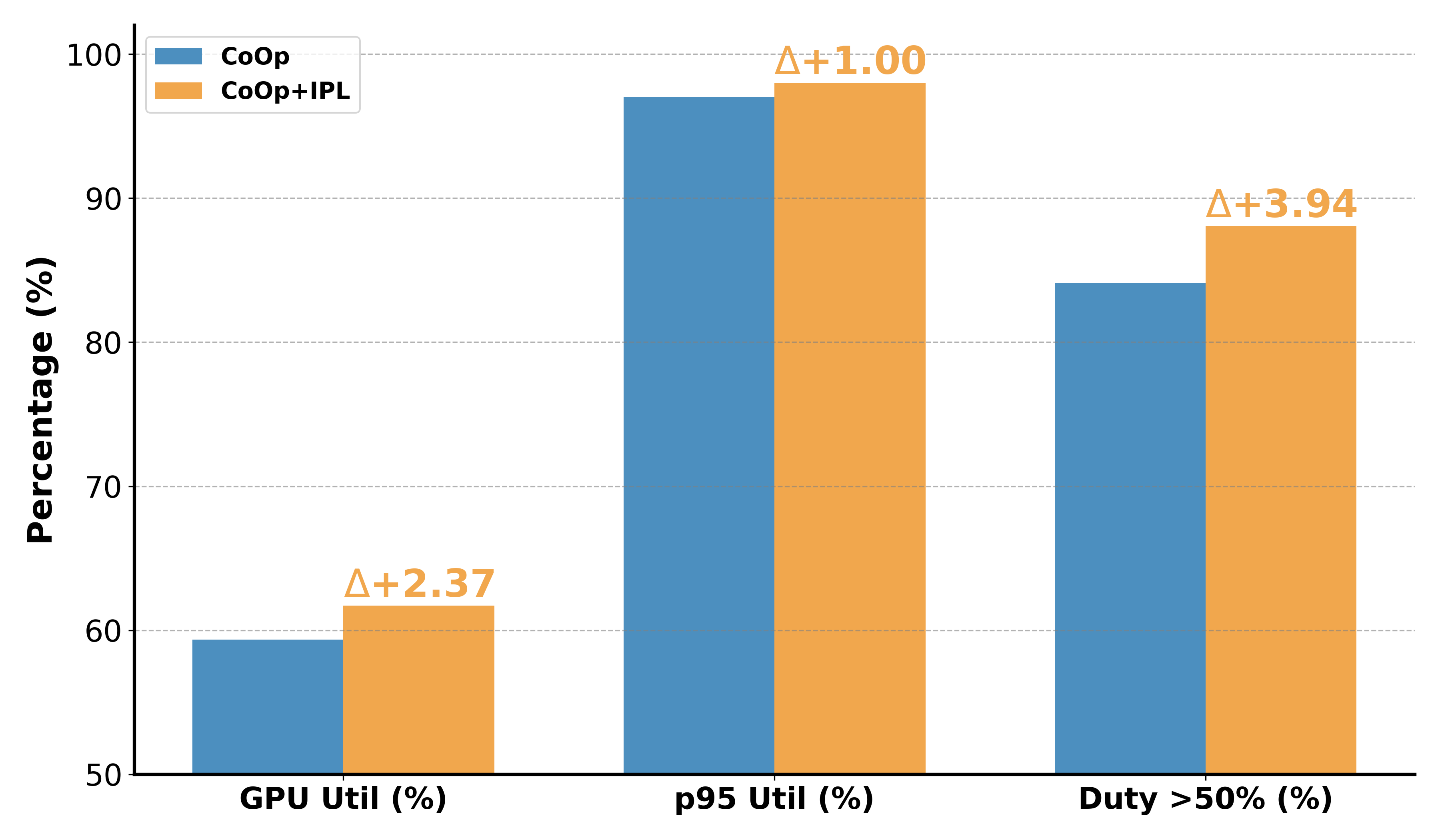}
    \caption{ Comparison of GPU utilization, p95 utilization, and duty cycle between CoOp and CoOp+IPL. CoOp+IPL introduces only modest increases in GPU usage across all three metrics.}
    \label{fig:bar}
\end{figure}

\subsubsection{Runtime and Overhead}
We evaluate the computational overhead introduced by the semantic token selection module in IPL. Across the 11 datasets used in our experiments, with a candidate vocabulary of 8,883 words, one full greedy traversal takes 252 seconds on average. Since each traversal evaluates candidate tokens in the same manner, the computational cost grows approximately linearly with the size of the candidate pool. In practice, this cost can be further reduced by restricting the pool to a more task-specific vocabulary.

The evaluation of candidate tokens is also amenable to parallelization. Specifically, if $B$ candidates are processed simultaneously, the wall-clock time can in principle be reduced by a factor close to $B$, subject to GPU memory and implementation overhead. Therefore, the reported runtime should be viewed as a conservative estimate under a standard sequential implementation, while additional acceleration is possible when hardware resources permit.

To further quantify the training overhead introduced by IPL, we monitor GPU behavior during training. As shown in Figure~\ref{fig:bar}, compared with CoOp, CoOp+IPL results in modest increases in three GPU metrics: average utilization, high-percentile utilization, and the duty ratio above 50\%. These metrics reflect overall load, peak demand, and the proportion of time the GPU remains actively engaged, respectively. The consistent trend across all three metrics indicates that incorporating IPL leads to only a limited increase in training-time GPU demand. In addition, the peak memory footprint increases from 3.69 GB to 4.66 GB, which remains modest and does not affect training feasibility on standard GPUs.

Overall, these results show that the additional cost of IPL is small relative to the full training pipeline, and that the greedy token selection stage remains lightweight in practice, even under shared or resource-constrained GPU environments.

\subsubsection{Selecting Templates}
To further study the effect of template design in Subsection~\ref{sec:selection}, we conduct an ablation study on three representative datasets using CoOp as the baseline. All variants are built upon the conventional prompt form ``a photo of a [CLS]'', with different extended templates used to introduce the selected tokens. Since different templates provide different contextual priors, they may lead to different token selections and thus different downstream performance. As shown in Table~\ref{tab:template_ablation}, the template ``with emphasis on'' consistently outperforms the other variants and yields the best harmonic mean accuracy on all three datasets. Based on this observation, we adopt ``a photo of a [CLS], with emphasis on: [...]'' as the default template in all experiments.

Beyond the final accuracy differences, template choice also influences the semantic tendency of token selection. Different templates reshape the contextual relationship between the class name and candidate words, thereby affecting their relative similarity ranking in the text embedding space. In our experiments, the template ``with emphasis on'' tends to favor more salient and dataset-relevant attributes, whereas other templates more often lead to generic or descriptive words. This suggests that template design affects not only classification performance but also the type of semantic features that are more likely to be selected.

\begin{table}[t]
    \centering
    \caption{Impact of Template Choice on Harmonic Mean Accuracy}
    \label{tab:template_ablation}
    \begin{tabular}{lccc}
        \toprule
        Template & Caltech101 & OxfordPets & DTD \\
        \midrule
        ``with emphasis on:'' & \textbf{95.19} & \textbf{95.99} & \textbf{63.30} \\
        ``focusing on:'' & 94.77 & 93.84 & 61.22 \\
        ``showcasing:'' & 94.50 & 95.76 & 60.11 \\
        ``with following aspects:'' & 95.08 & 94.15 & 62.96 \\
        \bottomrule
    \end{tabular}
\end{table}

\section{Conclusion and Future Work}
Adapting large-scale vision--language models to downstream tasks remains challenging due to overfitting under limited supervision and the lack of explicit semantics in learned prompts. Although prompt learning offers a parameter-efficient adaptation paradigm, the resulting prompts are often difficult to interpret, which can limit transferability and weaken generalization to novel categories.

To address these issues, we propose IPL, a prompt learning framework that embeds selected semantic tokens into continuous prompts, improving interpretability while alleviating overfitting to base classes. Extensive experiments on 11 datasets demonstrate consistent gains over strong baselines, including an average improvement of +1.26\% in harmonic mean for base-to-novel generalization. Additional improvements in cross-dataset and domain generalization further verify the robustness of our method. As a plug-and-play enhancement to existing prompt learning methods, IPL achieves better performance with only modest overhead, demonstrating both effectiveness and practicality.

Looking ahead, future work will investigate the incorporation of more dynamic and context-sensitive semantic priors to further enhance the flexibility and expressiveness of prompt representations. While our current design relies on handcrafted selection templates that perform well empirically, they remain static and lack adaptation to different contexts. In addition, we plan to extend our approach to broader modalities and domains, such as video-language understanding and medical imaging, where the dual objectives of interpretability and generalization remain paramount.

\bibliographystyle{cas-model2-names}
\bibliography{cas-refs}

\end{document}